\documentclass[letterpaper, 10 pt, conference]{ieeeconf}  % Comment this line out if you need a4paper

\IEEEoverridecommandlockouts                              % This command is only needed if 
\usepackage{microtype}
\usepackage{multirow}
\usepackage{graphicx}
\usepackage{diagbox}
\usepackage{amsmath}
\usepackage{amsfonts}
\usepackage{hyperref}
\usepackage{amssymb}
\usepackage{xcolor}

\title{\LARGE \bf
Robotic Occlusion Reasoning for Efficient Object Existence Prediction}

\author{Mengdi Li$^{1}$, Cornelius Weber$^{1}$, Matthias Kerzel$^{1}$, Jae Hee Lee$^{1}$, Zheni Zeng$^{2}$, Zhiyuan Liu$^{2}$, Stefan Wermter$^{1}$% <-this % stops a space
\thanks{$^{1}$Knowledge Technology, Department of Informatics, University of Hamburg, 22527 Hamburg, Germany.
        {\tt\small \{mli, weber, kerzel, lee, wermter\}@informatik.uni-hamburg.de}}%
\thanks{$^{2}$Department of Computer Science and Technology, Tsinghua University, 100084 Beijing, China.
        {\tt\small zzn20@mails.tsinghua.edu.cn, liuzy@tsinghua.edu.cn}}%
}

\begin{document}

\maketitle
\thispagestyle{empty}
\pagestyle{empty}

%%%%%%%%%%%%%%%%%%%%%%%%%%%%%%%%%%%%%%%%%%%%%%%%%%%%%%%%%%%%%%%%%%%%%%%%%%%%%%%%
\begin{abstract}
Reasoning about potential occlusions is essential for robots to efficiently predict whether an object exists in an environment. 
Though existing work shows that a robot with active perception can achieve various tasks, it is still unclear if occlusion reasoning can be achieved. 
To answer this question, we introduce the task of robotic object existence prediction: 
when being asked about an object, a robot needs to move as few steps as possible around a table with randomly placed objects to predict whether the queried object exists. 
To address this problem, we propose a novel recurrent neural network model that can be jointly trained with supervised and reinforcement learning methods using a curriculum training strategy. 
Experimental results show that 1) both active perception and occlusion reasoning are necessary to successfully achieve the task; 
2) the proposed model demonstrates a good occlusion reasoning ability by achieving a similar prediction accuracy to an exhaustive exploration baseline while requiring only about $10\%$ of the baseline's number of movement steps on average; and 3) the model generalizes to novel object combinations with a moderate loss of accuracy. 
\end{abstract}

%%%%%%%%%%%%%%%%%%%%%%%%%%%%%%%%%%%%%%%%%%%%%%%%%%%%%%%%%%%%%%%%%%%%%%%%%%%%%%%%
\section{INTRODUCTION}
Indoor assistant robots that are able to perform tasks, such as searching for objects and answering questions about the environment, according to verbal commands from users have promising application prospects. 
We expect robots to not only complete these tasks correctly but also complete them efficiently, which benefits improving user experience and reduces energy requirements. 

The ability of reasoning about potential occlusions of objects is essential for achieving the aforementioned goal. When asked to search for an object, a robot needs to reason whether the target object is possibly occluded by visible objects, and then determine whether to check the occluded space by executing movement actions. 
However, occlusion reasoning is non-trivial: a robot needs to know the size of the target object from the verbal instruction, and compare it with the size of the visible objects to perform occlusion reasoning. 
% Existing work
Though existing work has shown that robots with active perception can achieve various tasks \cite{zhu2017target, ye2018active, wang2018efficient, yang2019ear}, 
in this work we further investigate if robots can efficiently explore environments by performing occlusion reasoning. 

\begin{figure}
    \centering
    \includegraphics[scale=0.25]{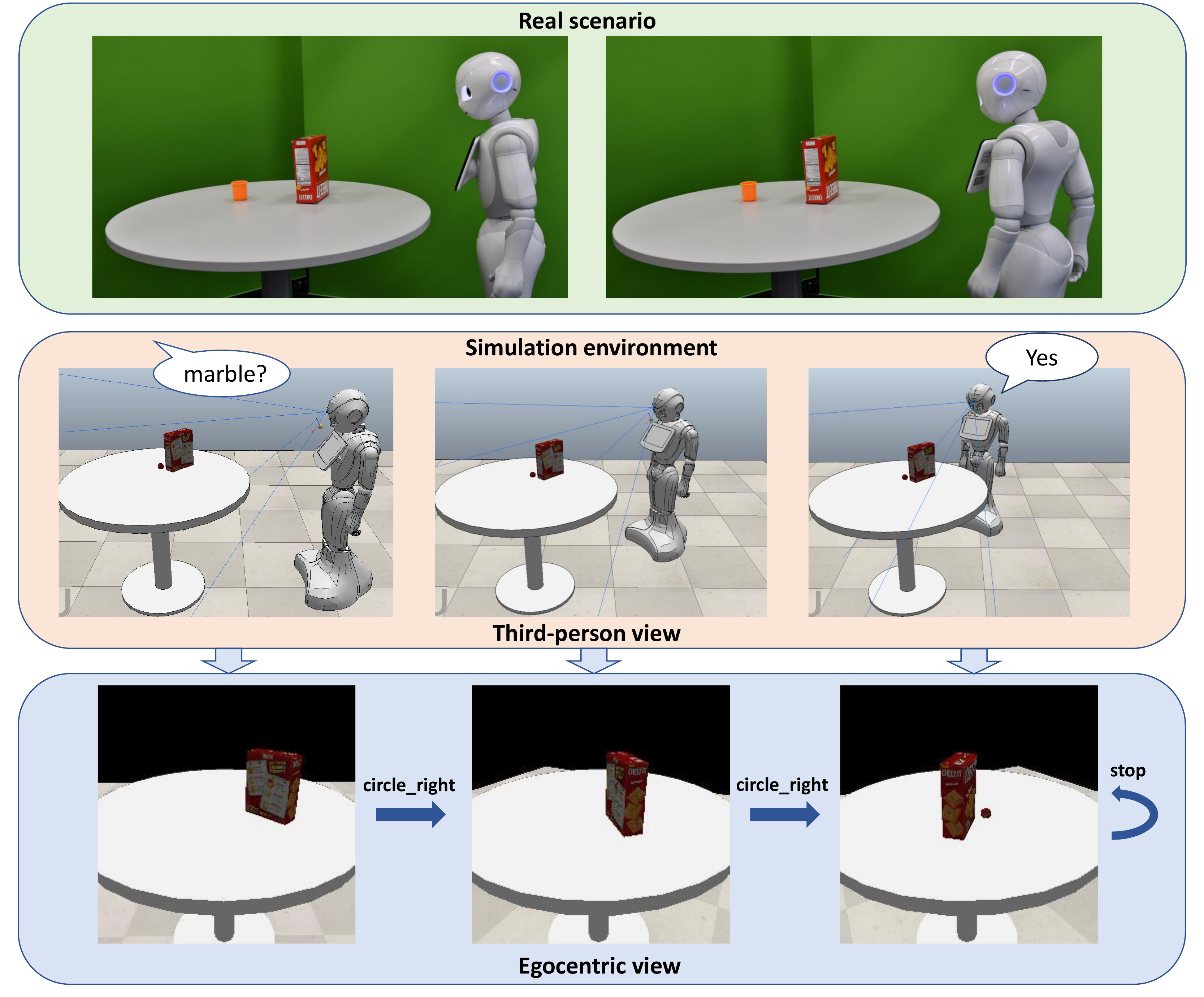}
    \caption{The task of robotic object existence prediction: given a word instruction (e.g. ``marble"), a robot standing by a table needs to execute as few movement steps as possible to give a correct prediction (e.g. yes) whether the queried object exists on the table.}
    \label{fig:environment}
\end{figure}

To answer this question, we propose a novel robotic object existence prediction (ROEP) task.  Fig.~\ref{fig:environment} shows the task in real scenarios, and a simulation environment built using the robot simulator CoppeliaSim~\cite{coppeliaSim}. 
The robot is the humanoid Pepper\footnote{\url{https://www.softbankrobotics.com/emea/en/pepper}.} from SoftBank Robotics, which has three omnidirectional wheels for flexible locomotion. 
The movement of the robot is implemented as a circular motion around the table by $30$ degrees clockwise or anticlockwise. 
The robot receives a word instruction (e.g. ``marble"), and is rewarded for correctly predicting whether the target object exists on the table while executing as few movement steps as possible. 
There are three main challenges behind achieving this goal: 
1) the robot needs to connect linguistic concepts with visual representations; 
2) the robot needs to memorize the past interactions with the environment to make action selection decisions; 
and 3) the selected actions and the final prediction functionally interact with each other, which makes the training difficult. 

\begin{figure*}
    \centering
    \includegraphics[scale=0.45]{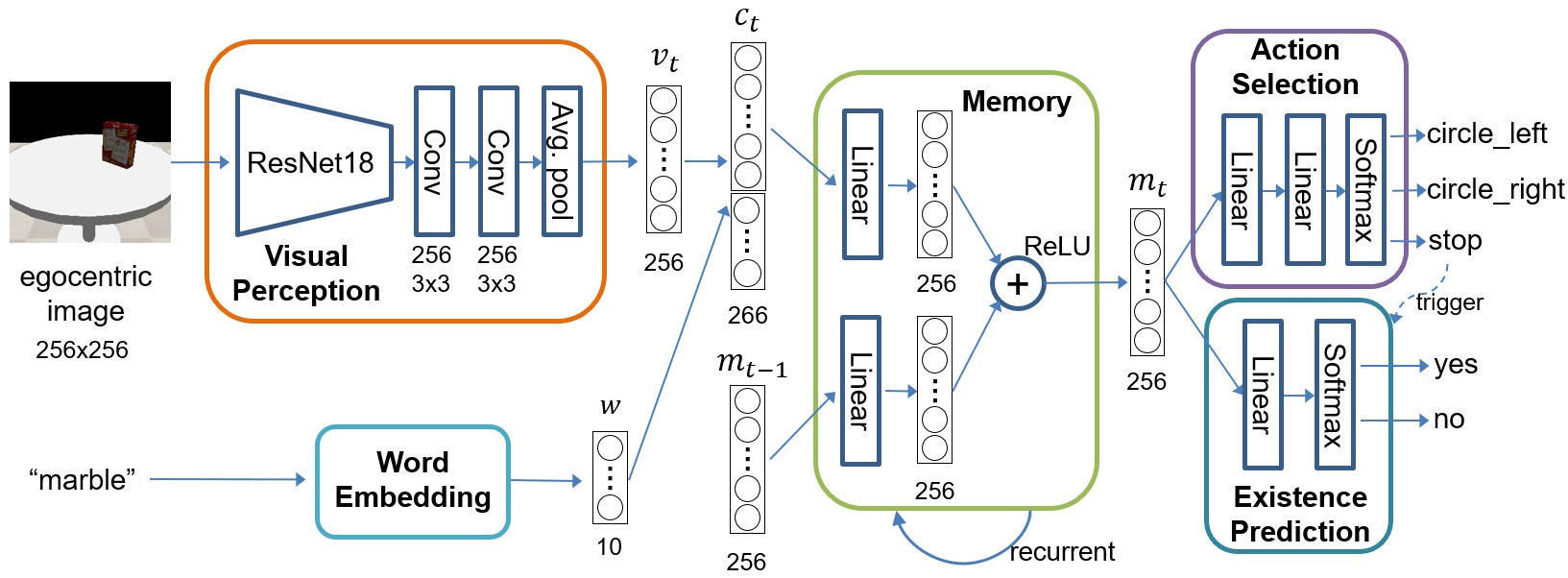}
    \caption{The architecture of the proposed model.}
    \label{fig_architecture}
\end{figure*}

% model
We propose a novel model (see Fig.~\ref{fig_architecture}) to address the above challenges. This model is a recurrent neural network consisting of five modules: a visual perception module, a word embedding module, a memory module, an action selection module, and an existence prediction module. 
The model can be jointly trained with reinforcement learning and supervised learning methods using a curriculum training strategy \cite{bengio2009curriculum}. 

% experiments, and findings
We evaluate our model by comparing it with three baselines, which are a passive model without any movement, a random model with a stochastic movement selection strategy, and an exhaustive exploration model which takes a maximum number of movements. 
Experimental results demonstrate that our model can outperform the passive and random baselines by a large margin, and achieve a similar prediction accuracy to the exhaustive exploration model while requiring only about $10\%$ of the baseline's number of movement steps on average. This shows the necessity of active perception and occlusion reasoning to successfully achieve the task, and that a good occlusion reasoning ability is obtained by our model. 

As the number of different objects increases, the number of possible combinations of two objects with occlusion increases exponentially. 
So a good generalization performance on novel combinations of two objects is especially important for occlusion reasoning. 
We evaluate the generalization performance of our model on novel object combinations held out from the training data, where we show that the generalization to novel object combinations comes with a moderate loss of accuracy while maintaining a small average number of movement steps. 
Moreover, the generalization performance increases when more kinds of object combinations are included in the training data. 

The main contributions of the paper can be summarized as follows: 1) we formulate a novel robotic object existence prediction (ROEP) task, which poses a high requirement of active perception and occlusion reasoning ability for robots; 2) we develop a novel model that can efficiently achieve this task; 
and 3) we find that the proposed model generalizes to novel object combinations with a moderate loss of accuracy, and that the variety of object combinations in the training data benefits increasing generalization. 

\section{RELATED WORK}
\textbf{Mobile robots with active perception: }
Zhu et al.~\cite{zhu2017target} proposed a reinforcement learning model for the task of target-driven visual navigation. The model is expected to navigate towards a visual target in indoor scenes with a minimum number of movement steps by its egocentric visual inputs and the image of the target. 
Ye et al.~\cite{ye2018active} studied the problem of mobile robots searching small target objects in arbitrary poses in indoor environments. They proposed a model integrating an object recognition module and a deep reinforcement learning-based action selection module together for the object searching task. 
Wang et al.~\cite{wang2018efficient} focused on the efficiency of robots when searching for target objects. They proposed a scheme to encode the prior knowledge of the relationship between rooms and objects in a belief map to facilitate efficient searching. 
Instead of focusing on achieving tasks in large-scale indoor environments, we concentrate on the efficiency of the model when encountering occlusion situations. 

\textbf{Object occlusion: }The occlusion situation between objects is very common in robotic scenarios. However, the occlusion reasoning ability of autonomous mobile robots has not been studied well. 
Yang et al.~\cite{yang2019ear} introduced the task of embodied amodal recognition focusing on the visual recognition ability of agents in scenes with occlusion. 
They proposed a model that can navigate in the environment to perform object classification, location, and segmentation. 
However, this work did not concentrate on the occlusion reasoning ability of the agent. 
% The mqa paper. 
A recent work on developing robots with the occlusion reasoning ability is \cite{deng2020mqa}. This work introduced the task of answering visual questions via manipulation (MQA), where a robot manipulator needs to perform a series of actions to move objects possibly occluding some small objects on a tabletop, in order to correctly answer visual questions. 
Similar to the ROEP task, the MQA task also requires the robot to have the ability of occlusion reasoning to perform reasonable exploration actions. However, the robot in MQA is a manipulator that explores the environment by moving objects, while we focus on autonomous mobile robots to explore the environment by active perception. 

\textbf{Embodied learning: }Robotics research is recently benefiting from achievements in vision and language processing. On the other hand, researchers are also taking advantage of agents situated in 3D environments to conduct multimodal research. 
It has been proven that an active agent is able to connect linguistic concepts with visual representations of the environment through training to complete action-involved tasks \cite{hermann2017grounded, chaplot2018gated}. 
Hill et al.~\cite{hill2020grounded} found that an embodied agent can achieve one-shot word learning when trained with reinforcement learning in a 3D environment. 
The proposed ROEP task also involves multimodalities, including vision, language, and action. 
Different from the abovementioned work, our model needs to specifically connect linguistic concepts with visual representations about object size through training to achieve the ROEP task. 

\begin{table}
\caption{Objects used in the simulation environment}\label{table:objs_table}
\centering
\begin{tabular}{|c|c c c c |}
\hline
Category & \multicolumn{4}{c|}{Objects} \\
\hline
\multirow{ 2}{*}{\textit{Large}} & cracker\_box & cleanser & laptop & pitcher \\
& desktop\_plant & wine & teddy\_bear & \\
\hline
\multirow{ 2}{*}{\textit{Medium}} & apple & baseball & foam\_brick & mug \\
& rubiks\_cube & meat\_can & coffee\_can & \\
\hline
\multirow{ 2}{*}{\textit{Small}} & bolt & dice & key & marble \\
& card & battery & button\_battery & \\
\hline
\end{tabular}
\end{table}

\section{Robotic Object Existence Prediction}

\subsection{Simulation Environment}
Existing simulation environments are not suitable for the ROEP task. We create a corresponding tabletop simulation environment using the robot simulator CoppeliaSim \cite{coppeliaSim} (see Fig.~\ref{fig:environment}). 
The robot can capture egocentric RGB images by a visual sensor mounted on its head, and execute actions selected from (\textit{circle\_left}, \textit{circle\_right}, and \textit{stop}). By taking the action \textit{circle\_left}, the robot circles around the table clockwise by 30 degrees. The action \textit{circle\_right} works in the same way but in an anticlockwise direction. When the action \textit{stop} is selected or the maximum number of $6$ movement steps is reached, the robot takes no movement action and predicts whether the queried object exists. 

A total of 21 everyday objects are used in the simulation environment. 
Some of them are from the YCB dataset \cite{calli2015benchmarking}. The rest of them are provided by CoppeliaSim or collected online. 
These objects are divided into 3 categories according to their relative size, as shown in Table~\ref{table:objs_table}. 
When fitting these objects into cubes, objects from the \textit{Large} category have a minimum height of $21cm$ and an average volume of $2905cm^3$. The heights of objects from the \textit{Medium} category are from $5cm$ to $14cm$, and their average volume is $508cm^3$. Objects from the \textit{Small} category have a maximum height of $3cm$ and an average volume of $7cm^3$. 
There are potential occlusions of objects from different categories. 

\subsection{Data Generation}
\label{subsec:data_generation}
Our data is automatically generated based on predefined rules like the CLEVR \cite{johnson2017clevr} and the ShapeWorld \cite{kuhnle2017shapeworld} datasets. 
All the samples are generated during training and testing periods. Each data sample is a triplet [\textit{Scene}, \textit{Query}, \textit{Prediction}]. \textit{Scene} is an arrangement of objects on the table. \textit{Query} is a word randomly selected with equal probability from Table~\ref{table:objs_table} to instruct the robot to search for the referred object in \textit{Scene}. 
\textit{Prediction} is a ground-truth binary label representing whether the target object exists in \textit{Scene}. It is randomly set as positive or negative with the equal probability of $50\%$. 
Based on a determined pair of \textit{Query} and \textit{Prediction}, a corresponding scene is then generated. 

There are three different types of scenes: 1) scenes that contain one object; 2) scenes with two objects without occlusion from the initial field of view of the robot; 3) scenes with two objects, one of which is occluded by the other one from the initial field of view of the robot. 
They account for the same proportion ($1\slash3$) in the generated data. 
To generate scenes with one object, the object is randomly placed on the table. 
To generate scenes with two objects, some geometric calculations using position coordinates of the robot's visual sensor, and both position coordinates and heights of the two objects are applied for controlling whether there is occlusions in generated scenes. 
It should be noted that the smaller object is not necessarily fully occluded by the larger one in scenes with occlusion. 

We have a reasoning table (see Table~\ref{table:reasoning_table}) of the ideal action strategy at the first time step in an episode. This table shows whether the robot should move to change its viewpoint or predict the existence of the target object directly when given a query for objects of a specific category (different columns), and the object seen from the initial viewpoint. 
Except for the situation where a \textit{Large} object is queried, or a \textit{Small} object is seen, the robot has to utilize both information from the word instruction and visual perception to make an ideal action decision. 
Because there are at most two objects on the table, whenever the robot sees two objects, the robot should give an existence prediction directly no matter which object is queried.

\begin{table}
\caption{Reasoning table}\label{table:reasoning_table}
\centering
\begin{tabular}{|l|*{3}{c c c|}}\hline
& \multicolumn{3}{c|}{Query}\\\hline
Visible Object & \textit{Large} & \textit{Medium} & \textit{Small}\\\hline
One \textit{Large} & predict & move & move\\
One \textit{Medium} & predict & predict & move\\
One \textit{Small} & predict & predict & predict\\\hline
\end{tabular}
\end{table}

\section{METHODOLOGY}

\subsection{Model}
Our proposed model is inspired by the recurrent attention model~\cite{mnih2014recurrent}, which is originally applied to attention-driven image classification tasks. 
The proposed model is a recurrent neural network overall (see Fig.~\ref{fig_architecture}), and can be divided into five parts: 1) a memory module for incrementally building up state representations, 2) a visual perception module for extracting visual representations, 3) a word embedding module for extracting distributed representations of a query word, 4) an action selection module for making action decisions, and 5) an existence prediction module for producing final predictions.

The \textbf{Visual Perception} module takes the egocentric RGB image ($256\times256$ pixels) as input to extract visual representations. It first extracts the 128 28$\times$28 image feature maps from the \textit{conv3} layer of a fixed ResNet18 \cite{he2016deep} pretrained on ImageNet \cite{russakovsky2015imagenet}. The feature is then passed through two CNN layers both with 256 3$\times$3 kernels, and an average pooling layer to obtain the visual representations $v_t$ with a length of $256$. This process is similar to the visual module of the MAC model \cite{hudson2018compositional} designed for visual reasoning on the CLEVR dataset~\cite{johnson2017clevr}. 

The \textbf{Word Embedding} module maps each word instruction to a 10-dimensional word vector $w$. 
The weights of the embedding module are randomly initialized, and updated during training. 

The \textbf{Memory} module is a recurrent unit that takes the concatenated representations $c_t = (v_t, w)$ as the input, and combines $c_t$ with the internal representations at the previous time step $m_{t-1}$ to produce the new internal representations $m_t\in \mathbb{R}^{256\times1}$. This process can be formalized as
\iffalse
\begin{equation} \label{eqn:rnn_function}
\begin{split}
m_t & = f_m(m_{t-1}, c_t)\\
    & = \text{ReLU}((W_m \cdot m_{t-1} + b_m) + (W_c \cdot c_t + b_c))
\end{split}
\end{equation}
\fi
\begin{equation} \label{eqn:rnn_function}
m_t = f_m(m_{t-1}, c_t) = \text{ReLU}(W_m \cdot m_{t-1} + W_c \cdot c_t + b)
\end{equation}
where $W_m \in \mathbb{R}^{256\times256}$ and $W_c \in \mathbb{R}^{256\times266}$ are weight matrices, $b \in \mathbb{R}^{256\times1}$ is a bias vector, ReLU($\cdot$) is the rectified linear activation function. 
More sophisticated units such as LSTM or GRU are not used for the memory module because a vanishing gradient is not a problem for our task since only few recurrent steps have to be taken. 

The \textbf{Action Selection} module and \textbf{Existence Prediction} module are both classification networks with \textit{softmax} outputs. 
The action selection module is a fully connected network with one hidden layer ($128$ hidden units). Its three \textit{softmax} outputs correspond to three movement actions. 
The existence prediction module has a single linear layer followed by a \textit{softmax} layer with two outputs which correspond to the positive and negative prediction respectively. 

\subsection{Training}
The parameters of our model include parameters of the visual perception module, the word embedding module, the memory module, the action selection module, and the existence prediction module $\theta = \{\theta_v, \theta_w, \theta_m, \theta_{a}, \theta_{p}\}$. The model is non-differential overall. We train the model jointly with supervised learning and reinforcement learning methods, where $\theta_{a}$ is trained using reinforcement learning, $\{\theta_v, \theta_w, \theta_m, \theta_{p}\}$ are trained using supervised learning. 

% formulate the task with RL concepts
The task can be formalized as a partially observable Markov decision process from the perspective of reinforcement learning. The true state of the environment cannot be fully observed. The action selection module is a reinforcement learning agent, which needs to learn a stochastic policy $\pi(a_t|s_{0:t};\theta_{a})$ with the parameters $\theta_{a}$, where $a_t$ is one of the three actions in the predefined action set. Executing each movement action except the \textit{stop} action leads the model to obtain a new visual input. $s_{0:t} = w, v_0, a_0, v_1, a_1, ..., v_t$ is the history of past interactions with the environment from time step $0$ to $t$. The internal representations $m_t$ in the memory module is an approximation to $s_{0:t}$. 

% reward define
The model is expected to gain a high reward at the end of each episode. 
We design a cost-sensitive reward function containing two parts, an accuracy reward $r_{acc}$ and a latency reward $r_{lat}$. An accuracy reward of $1$ is received when a correct prediction is produced. An accuracy reward of $-1$ is received when an incorrect prediction is produced. The latency reward is
\begin{equation} \label{eqn:latency_reward}
r_{lat} = \frac{1}{T + 2}
\end{equation}
where $T$ is the number of movement steps the agent takes in one episode. $T = 0$ means that the \textit{stop} action is selected at time step $0$. 
The total reward at time step $T$ is a summation of these two rewards: $r_T = r_{acc} + r_{lat}$. 
We use $T + 2$ rather than $T + 1$ as the denominator of $r_{lat}$ to make sure that $r_T$ is negative when the prediction is incorrect. 
At other time steps ($t = 0,...,T-1$), we set $r_t = 0$. 

The agent is expected to maximize the expected reward return $J(\theta_{a})$ under the policy $\pi(a_t|s_{0:t};\theta_{a})$. 

\begin{equation} \label{eqn:reward_function}
J(\theta_{a}) = \mathbb{E}_{\pi(a_t|s_{0:t};\theta_{a})}\left[\sum_{t=0}^{T} r_t\right]
\end{equation}

We use Monte-Carlo policy gradient (REINFORCE) \cite{williams1992simple} to optimize the agent. 
REINFORCE uses the sample gradient to approximate the actual gradient of $J(\theta_{a})$

\begin{equation} \label{eqn:adjusted_reward_gradient_function}
\nabla_{\theta_{a}}J \approx \sum_{t=0}^{T} \nabla_{\theta_{a}} \log\pi(a_t|s_{0:t};\theta_{a})(R_t-b_t)
\end{equation}
where $R_t = \sum_{t^\prime=0}^{T}r_{t^\prime}$ is the accumulated reward following the action $a_t$, $b_t$ is the estimated reward predicted by a baseline network, which has a single linear layer taking $m_t$ as the input. The estimated reward $b_t$ is used for reducing variance of gradient estimation. 
The baseline network is trained with a mean squared error loss $\mathcal{L}_{b} = \frac{1}{T}\sum_{t=0}^T(R_t-b_t)$. 

To use gradient descent algorithms for optimizing the agent, we define loss $\mathcal{L}_{a} = - J(\theta_{a})$. It should be noted that gradients of $\mathcal{L}_{a}$ and $\mathcal{L}_{b}$ are not backpropagated to the memory, visual perception, and word embedding module. 

% supervised training part
We train these modules along with the existence prediction module using supervised learning methods to optimize the binary cross-entropy loss
\begin{equation} \label{eqn:loss_ans}
\mathcal{L}_{p} = - y\log\hat{y} - (1-y)\log(1-\hat{y})
\end{equation}
where $y$ is the labeled ground-truth prediction ($1$ for yes, $0$ for no), $\hat{y}$ is the estimated probability of the prediction yes. Gradients of $\mathcal{L}_{p}$ are backpropagated to update parameters of the existence prediction $\theta_{p}$, memory $\theta_m$, visual perception $\theta_v$, and word embedding $\theta_w$ module. 

% total loss
The total loss function is a weighted summation of the three losses, as
\begin{equation} \label{eqn:loss_total}
\mathcal{L}_{total} = \mathcal{L}_{p} + \alpha \cdot \mathcal{L}_{a} + \beta \cdot \mathcal{L}_{b}
\end{equation}
where $\alpha$ and $\beta$ are weight coefficients of $\mathcal{L}_a$ and $\mathcal{L}_{b}$ respectively.

\subsection{Training Details}
We found that it is hard to train the model from scratch on data with all three different types of scenes, which corresponds to the finding of \cite{yang2019ear} that joint training perception and policy networks from scratch is difficult. 
We resort to a curriculum training strategy to train the model on data with 4 levels of increasing difficulty. 
We refer to data only containing scenes with one object as \textit{L1-1-vis}, data only containing scenes with two objects without occlusion as \textit{L2-2-vis}, data only containing scenes with two objects with occlusion as \textit{L3-2-occ}, and data containing all types of scenes as \textit{L4-overall}, in which three types of scenes occupy the same proportion. 
The model is trained on these four levels of data sequentially. The parameters obtained from one training stage are loaded as the initial parameters for the next training stage. 

We use the Adam optimizer with a learning rate of $1e-4$. The weight coefficients in the total loss function (Eq.~\ref{eqn:loss_total}) is set as $\alpha = 1e-2$, $\beta = 1$ for training stage on the first three levels. 
A smaller weight coefficient $\alpha = 1e-4$ is used for the last training level to make the training process stabler.

\section{EXPERIMENTS}

\subsection{Curriculum Training}
Our model is trained using a curriculum training strategy. Specifically, the model is trained sequentially on \textit{L1-1-vis}, \textit{L2-2-vis}, \textit{L3-2-occ}, and \textit{L4-overall} data with a fixed number of episodes ($900k$, $900k$, $400k$, and $400k$ respectively) in our experiments. The total training process takes about four days using one GPU (NVIDIA Titan RTX). We noticed that it is unnecessary to train the model to achieve the best performance in the first three training stages if we are only interested in the final model. 
We repeat the experiment three times to avoid the effect of randomness. 
The accuracy of correct predictions and the average number of movement steps are used as metrics to evaluate the performance. 

\begin{figure}
    \centering
    \includegraphics[scale=0.37]{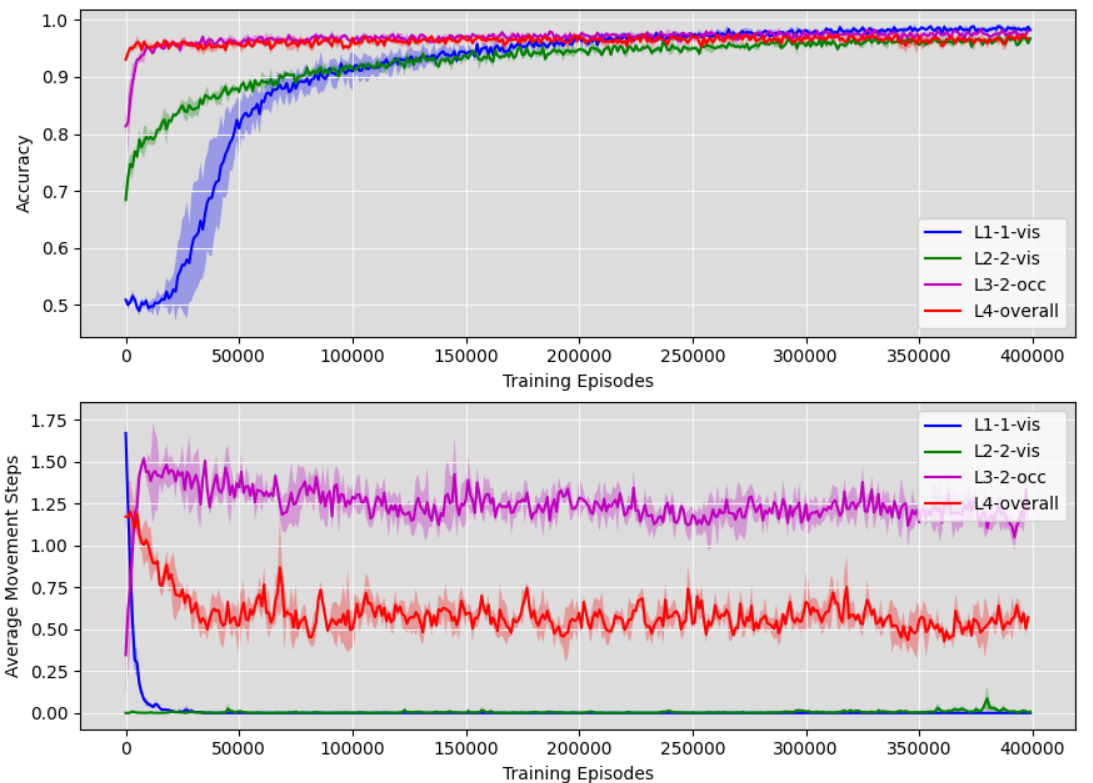}
    \caption{Training curves of the proposed model in different training stages. The model is sequentially trained on \textit{L1-1-vis}, \textit{L2-2-vis}, \textit{L3-2-occ}, and \textit{L4-overall} data. The parameters obtained from one training stage are loaded as the initial parameters for the next training stage.} 
    \label{fig:full_model_training_curves}
\end{figure}

Fig.~\ref{fig:full_model_training_curves} shows the training curves in different training stages. 
In the first two training stages on \textit{L1-1-vis} and \textit{L2-2-vis} data, the accuracy increases stably until reaching a plateau of over $97\%$, while the average number of movement steps stay near $0$. 
In the third training stage on \textit{L3-2-occ} data, the accuracy rapidly increases in the first $30k$ episodes with the rapid increase of the average movement steps.
In the last two training stages on \textit{L3-2-occ} and \textit{L4-overall} data, the average movement steps continuously decrease after the accuracy has reached a plateau. 

\begin{figure*}
    \centering
    \includegraphics[scale=0.5]{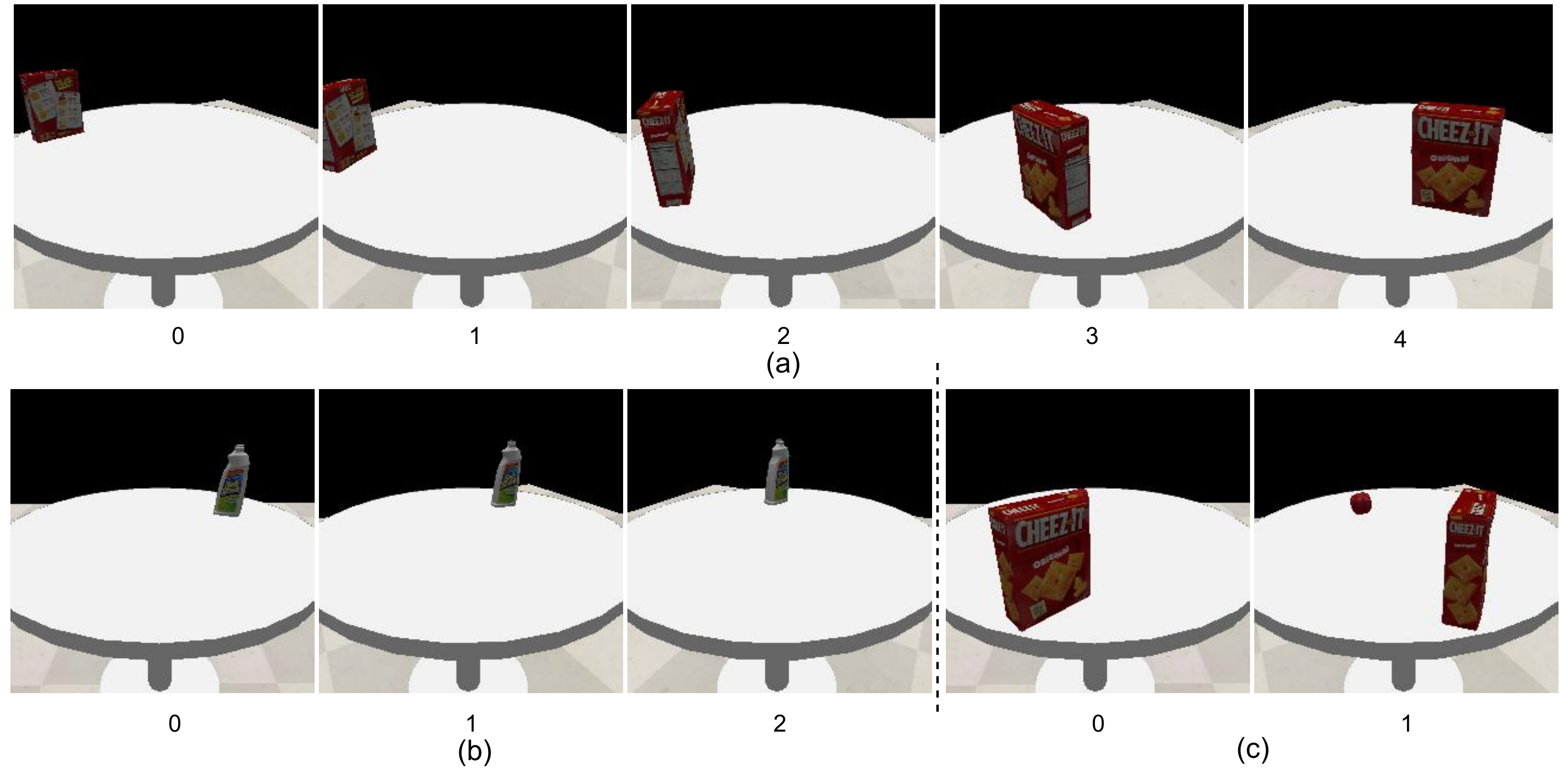}
    \caption{Examples of egocentric images in an episode when the query is ``apple". Numbers below the images indicate the time steps in an episode. 
    (a), (b): only one object larger than the target object exists; (c): the target object is occluded by a larger object. 
    In all cases, the agent moves to check the occluded space and provides the correct answer after the last shown frame.}
    \label{case_study_1}
\end{figure*}

We refer to models obtained from the first three training stages at the $900k$, $900k$, $400k$ episodes as \textit{Model$_{L1}$}, \textit{Model$_{L2}$}, \textit{Model$_{L3}$} respectively. 
The final model is obtained from the last training stage at the $400k$ episodes, and denoted as \textit{Final Model}. 
Performance of each model when tested on different test data ($10k$ episodes) is presented in Table~\ref{table:test_results}. 
The results show that each model scores well on the test data that corresponds to the training statistics (diagonal in bold font), and that the final model performs nearly as well as the individual models on their test data. 
% Moreover, the final model moves also when there is only one object, which makes sense because the model needs to check the occluded space in some cases. 
Fig.~\ref{case_study_1} shows examples when there is only one object, which is larger than the target object, visible from the initial perspective of the agent. 
A video showing the experimental results is available at \url{https://youtu.be/L4p7yo8dMmQ}. 
% \footnote{A video showing the experimental results is available at \url{https://youtu.be/L4p7yo8dMmQ}.}. 
% Footnote: A video showing the experimental results is available at xxx

\begin{table}
\caption{Performance evaluation on different test data}\label{table:test_results}
\centering
\scriptsize
\begin{tabular}{|l|*{4}{p{0.45cm} p{0.4cm}|}}
\hline
& \multicolumn{2}{c|}{$Model_{L1}$} & \multicolumn{2}{c|}{$Model_{L2}$} & \multicolumn{2}{c|}{$Model_{L3}$} & \multicolumn{2}{c|}{$Final\ Model$}\\
\hline
Test Data & Acc. & Steps & Acc. & Steps & Acc. & Steps & Acc. & Steps\\
\hline
\textit{L1-1-vis} & \textbf{99.4}\% & 0.0 & 90.9\% & 0.02 & 88.7\% & 1.20 & 99.0\% & 0.74\\
\textit{L2-2-vis} & 68.3\% & 0.0 & \textbf{97.4\%} & 0.01 & 91.2\% & 0.64 & 97.1\% & 0.39\\
\textit{L3-2-occ} & 74.4\% & 0.0 & 77.7\% & 0.07 & \textbf{98.3\%} & 1.23 & 96.9\% & 1.02\\
\textit{L4-overall} & 80.8\% & 0.0 & 88.5\% & 0.03 & 92.6\% & 1.00 & \textbf{97.2\%} & 0.71\\
\hline
\end{tabular}
\end{table}

\subsection{Comparison with Baselines}
We compare the proposed model with three baselines that have the same architecture as the proposed model, but with different action selection strategies. These baselines include a passive model without any movement, a random model with a stochastic movement selection strategy, and an exhaustive exploration model that executes the \textit{circle\_left} action for a maximum number of movement steps before producing a prediction. 
The average movement steps of the three baselines are $0$, $1.82$, and $6$ respectively. 

\begin{figure}
    \centering
    \includegraphics[scale=0.43]{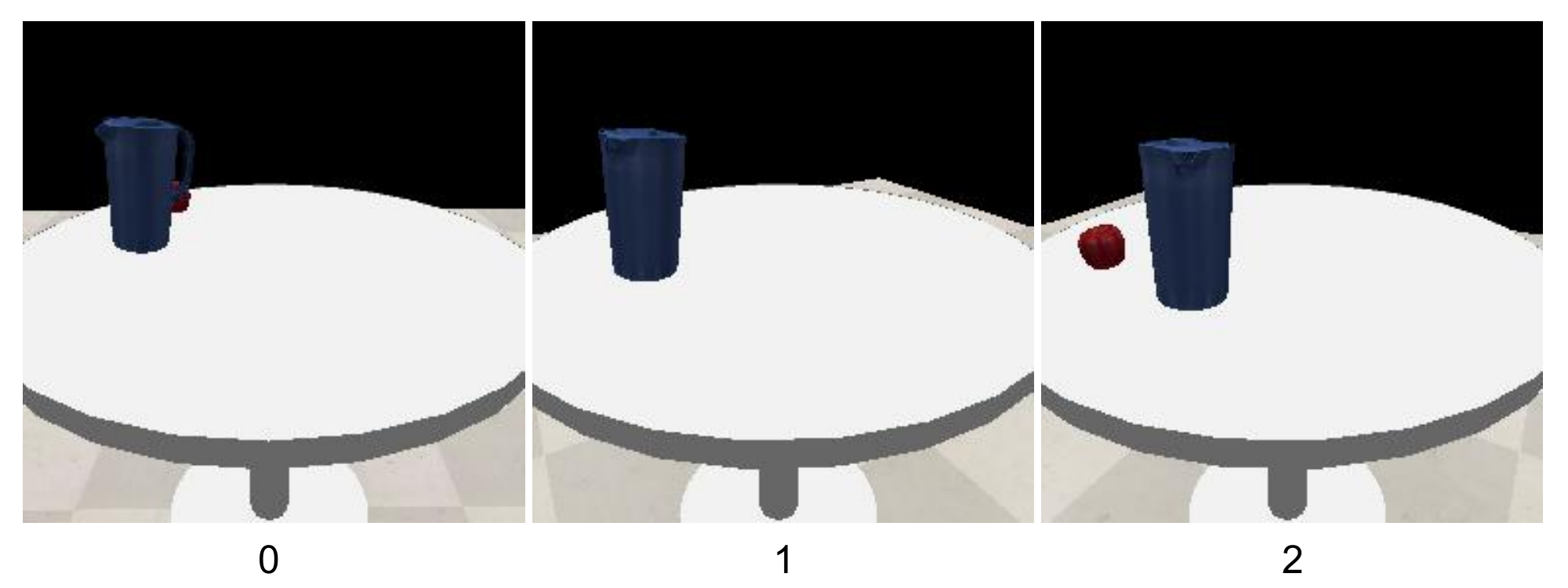}
    \caption{In this example, an apple is partially occluded by a pitcher. When the query is ``apple", the agent does not choose the optimal action \textit{circle\_right}, instead it chooses the action \textit{circle\_left}.}
    \label{case_study_2}
\end{figure}

The prediction accuracy of these baselines and our final model when tested on different test data is presented in Table~\ref{table:baselines_comparision}. The passive model and the random model are able to achieve a performance close to that of the exhaustive model on \textit{L1-1-vis}, and \textit{L2-2-vis} data, but performs poorly on \textit{L3-2-occ} data. 
This reveals that active perception is necessary to address the ROEP task. 
Our model can achieve a similar accuracy on all test data to the exhaustive model while requiring only $11.8\%$ of the baseline's number of movement steps on average ($0.71$ steps by our model, $6$ steps by the exhaustive model). 
This demonstrates that our model has obtained a good occlusion reasoning ability. However, there are still some challenges remaining: 1) The model learns to always choose one direction to move, rather than choose the optimal direction according to the orientation of the visible object or partial occlusion to check the occluded space (see Fig.~\ref{case_study_2}); 2) The model moves $0.39$ steps on average in scenes without occlusion (\textit{L2-2-vis}), which is unnecessary.

\begin{table}
\caption{Performance comparison with baselines}\label{table:baselines_comparision}
\centering
\scriptsize
\setlength\tabcolsep{4pt} % default value: 6pt
\begin{tabular}{|l|*{4}{c}|}
\hline
Test Data & Passive Model & Random Model & Exhaustive Model & Our Model\\
\hline
\textit{L1-1-vis} & 98.0\% & 97.6\% & \textbf{99.2\%} & 99.0\% \\
\textit{L2-2-vis} & 96.1\% & 92.6\% & 96.4\% & \textbf{97.1\%} \\
\textit{L3-2-occ} & 77.6\% & 83.3\% & \textbf{97.4}\% & 96.9\% \\
\textit{L4-overall} & 90.3\% & 91.1\% & \textbf{97.4\%} & 97.2\% \\
\hline
\end{tabular}
\end{table}

\subsection{Generalization Evaluation}

A good generalization performance on novel combinations of two objects is especially important for occlusion reasoning. 
To evaluate the generalization performance, we train our network on two different sets of training data excluding some object combinations, which are called holdout combinations. 
That means scenes with some specific object combinations, e.g. [mug, battery], are not included in the training data. 

There are three types of combinations of two different size categories, namely [\textit{Large}, \textit{Medium}], [\textit{Large}, \textit{Small}], [\textit{Medium}, \textit{Small}], and 147 $(7 \times 7 \times 3)$ possible combinations of two objects from different size categories. 
In the first training set, 21 object combinations (7 for each category combination) are held out only for testing, which accounts for $14.3\%$ of all possible combinations. 
In the second set, 42 object combinations (14 for each category combination) are held out, which accounts for $28.6\%$ of all possible combinations. 
Holdout combinations are determined by randomly selecting from all possible object combinations before the start of training. Every object in Table~\ref{table:objs_table} is shown in the training data. Experiments are repeated three times with different holdout combinations and random initialization. 

Table~\ref{table:test_holdout_results} presents the test results of the models trained on aforementioned two sets of training data, denoted as \textit{21 holdout} and \textit{42 holdout} respectively. 
Test data \textit{L2-2-vis (training)} and \textit{L3-2-occ (training)} contain scenes with object combinations used for training. Test data \textit{L2-2-vis (holdout)} and \textit{L3-2-occ (holdout)} only contain scenes with holdout object combinations. 
The results show that the two models can achieve similar high performance on scenes with object combinations used for training. 
When tested on \textit{L2-2-vis (holdout)} and \textit{L3-2-occ (holdout)}, the model trained on \textit{21 holdout} can still work well with an accuracy of over $90\%$ and a small average number of movement steps. The performance of the model trained on \textit{42 holdout} drops moderately to $86.7\%$ accuracy when tested on \textit{L3-2-occ (holdout)}, where occlusion reasoning on novel combinations is necessary.

\begin{table}
\caption{Generalization Evaluation}\label{table:test_holdout_results}
\centering
\begin{tabular}{|l|*{4}{c|}}
\hline
& \multicolumn{2}{c|}{\textit{21 holdout}} & \multicolumn{2}{c|}{\textit{42 holdout}}\\
\hline
Test Data & Acc. & Steps & Acc. & Steps\\
\hline
\textit{L1-1-vis} & 98.6\% & 0.651 & 98.8\% & 0.629\\
\textit{L2-2-vis (training)} & 96.8\% & 0.262 & 97.2\% & 0.197\\
\textit{L3-2-occ (training)} & 94.7\% & 0.873 & 96.2\% & 0.892\\
\textit{L2-2-vis (holdout)} & 95.8\% & 0.399 & 92.7\% & 0.317\\
\textit{L3-2-occ (holdout)} & 91.5\% & 0.895 & 86.7\% & 0.841\\
\hline
\end{tabular}
\end{table}

\section{Discussion}
\textbf{Experimental setup: }
The current experimental setup is simplified. There is a strong prior that there are at most two objects existing on the table, which limits the complexity of potential occlusion situations. 
An interesting extension is to generate scenes with more objects on the table and extend the task to counting objects. 
Moreover, the action space of the robot is small. The actions of \textit{circle\_left} and \textit{circle\_right} used in the current experimental setting limits the generalization capability to environments with tables of different sizes or shapes. 
More complex robot actions also involving \textit{move\_ahead}, \textit{rotate\_left}, \textit{rotate\_right}, etc., will be used in future work. On the one hand, it will be feasible to transfer a robot with these more complex actions to other environments. On the other hand, it will make the task more challenging, as the robot has greater flexibility in its movements, which places higher demands on action planning. 

\textbf{Training complexity: }The current training process is complex, since the curriculum training strategy involves four sequential training stages to obtain the final model. 
A possible solution to simplify training is using unsupervised learning \cite{ha2018world} instead of curriculum learning to learn good visual and word representations. 

\textbf{Sim-to-real transfer: }In this paper, we validate the effectiveness of the proposed model in a simulation environment. 
We can imagine that directly transferring the resulting model trained in a simulation environment to a real-world scenario (see Fig.~\ref{fig:environment}) would result in a certain performance loss. 
Some techniques, such as fine-tuning the model in a more photo-realistic simulation environment with randomized lighting conditions of the real environment, may mitigate the performance degradation. 

\section{CONCLUSION}
In this work, we introduced the task of robotic object existence prediction (ROEP), which is complementary to existing robotic tasks that require active perception. Different from existing tasks, ROEP focuses on the occlusion reasoning ability, which helps a robot explore its environments more efficiently. As such, it can be used for example as a probing task for existing models that are designed for general tasks. 

To solve ROEP we proposed a novel recurrent neural network which is trained end-to-end jointly with reinforcement learning and supervised learning methods using a curriculum training strategy. We showed empirically that the proposed model can efficiently achieve the ROEP task compared with the baselines. We also showed that generalization to novel object combinations comes with a moderate loss of accuracy, while including more kinds of object combinations in the training data can increase the generalization performance. This finding, which is related to the finding in~\cite{hill_environmental_2020}, can be considered as a recommendation when training a model for tasks that implicitly involve occlusion reasoning (e.g., object goal navigation \cite{chaplot2020object}). 
% active object searching or embodied question answering \cite{das2018embodied}). 

\textbf{Acknowledgement.} The authors gratefully acknowledge support from the China Scholarship Council (CSC) and the German Research Foundation DFG under project CML (TRR 169). We would like to thank Erik Strahl for his support with the experimental setup.

\addtolength{\textheight}{-12cm}   % This command serves to balance the column lengths

%%%%%%%%%%%%%%%%%%%%%%%%%%%%%%%%%%%%%%%%%%%%%%%%%%%%%%%%%%%%%%%%%%%%%%%%%%%%%%%%

%%%%%%%%%%%%%%%%%%%%%%%%%%%%%%%%%%%%%%%%%%%%%%%%%%%%%%%%%%%%%%%%%%%%%%%%%%%%%%%%



%\section*{ACKNOWLEDGMENT}
%The authors gratefully acknowledge support from the China Scholarship Council (CSC) and the German Research Foundation DFG under project CML (TRR 169). 

%%%%%%%%%%%%%%%%%%%%%%%%%%%%%%%%%%%%%%%%%%%%%%%%%%%%%%%%%%%%%%%%%%%%%%%%%%%%%%%%

\bibliographystyle{IEEEtran}
\bibliography{mybibliography}

\begin{thebibliography}{10}
\providecommand{\url}[1]{#1}
\csname url@rmstyle\endcsname
\providecommand{\newblock}{\relax}
\providecommand{\bibinfo}[2]{#2}
\providecommand\BIBentrySTDinterwordspacing{\spaceskip=0pt\relax}
\providecommand\BIBentryALTinterwordstretchfactor{4}
\providecommand\BIBentryALTinterwordspacing{\spaceskip=\fontdimen2\font plus
\BIBentryALTinterwordstretchfactor\fontdimen3\font minus
  \fontdimen4\font\relax}
\providecommand\BIBforeignlanguage[2]{{%
\expandafter\ifx\csname l@#1\endcsname\relax
\typeout{** WARNING: IEEEtran.bst: No hyphenation pattern has been}%
\typeout{** loaded for the language `#1'. Using the pattern for}%
\typeout{** the default language instead.}%
\else
\language=\csname l@#1\endcsname
\fi
#2}}

\bibitem{zhu2017target}
Y.~Zhu, R.~Mottaghi, E.~Kolve, J.~J. Lim, A.~Gupta, L.~Fei-Fei, and A.~Farhadi,
  ``Target-driven visual navigation in indoor scenes using deep reinforcement
  learning,'' in \emph{2017 IEEE International Conference on Robotics and
  Automation (ICRA)}, 2017, pp. 3357--3364.

\bibitem{ye2018active}
X.~Ye, Z.~Lin, H.~Li, S.~Zheng, and Y.~Yang, ``Active object perceiver:
  Recognition-guided policy learning for object searching on mobile robots,''
  in \emph{2018 IEEE/RSJ International Conference on Intelligent Robots and
  Systems (IROS)}, 2018, pp. 6857--6863.

\bibitem{wang2018efficient}
C.~Wang, J.~Cheng, J.~Wang, X.~Li, and M.~Q.-H. Meng, ``Efficient object search
  with belief road map using mobile robot,'' \emph{IEEE Robotics and Automation
  Letters}, vol.~3, no.~4, pp. 3081--3088, 2018.

\bibitem{yang2019ear}
J.~{Yang}, Z.~{Ren}, M.~{Xu}, X.~{Chen}, D.~{Crandall}, D.~{Parikh}, and
  D.~{Batra}, ``Embodied amodal recognition: Learning to move to perceive
  objects,'' in \emph{2019 IEEE/CVF International Conference on Computer Vision
  (ICCV)}, 2019, pp. 2040--2050.

\bibitem{coppeliaSim}
E.~Rohmer, S.~P.~N. Singh, and M.~Freese, ``{CoppeliaSim} (formerly {V-REP}): a
  versatile and scalable robot simulation framework,'' in \emph{Proceedings of
  The International Conference on Intelligent Robots and Systems (IROS)}, 2013,
  www.coppeliarobotics.com.

\bibitem{bengio2009curriculum}
Y.~Bengio, J.~Louradour, R.~Collobert, and J.~Weston, ``Curriculum learning,''
  in \emph{Proceedings of the 26th Annual International Conference on Machine
  Learning (ICML)}, 2009, pp. 41--48.

\bibitem{deng2020mqa}
Y.~Deng, D.~Guo, X.~Guo, N.~Zhang, H.~Liu, and F.~Sun, ``{MQA}: Answering the
  question via robotic manipulation,'' in \emph{Proceedings of Robotics:
  Science and Systems (RSS)}, 2021.

\bibitem{hermann2017grounded}
K.~M. Hermann, F.~Hill, S.~Green, F.~Wang, R.~Faulkner, H.~Soyer,
  D.~Szepesvari, W.~M. Czarnecki, M.~Jaderberg, D.~Teplyashin, \emph{et~al.},
  ``Grounded language learning in a simulated 3{D} world,'' \emph{arXiv
  preprint arXiv:1706.06551}, 2017.

\bibitem{chaplot2018gated}
D.~S. Chaplot, K.~M. Sathyendra, R.~K. Pasumarthi, D.~Rajagopal, and
  R.~Salakhutdinov, ``Gated-attention architectures for task-oriented language
  grounding,'' in \emph{Proceedings of the AAAI Conference on Artificial
  Intelligence (AAAI)}, vol.~32, no.~1, 2018.

\bibitem{hill2020grounded}
F.~Hill, O.~Tieleman, T.~von Glehn, N.~Wong, H.~Merzic, and S.~Clark,
  ``Grounded language learning fast and slow,'' in \emph{Proceedings of the
  International Conference on Learning Representations (ICLR)}, 2021.

\bibitem{calli2015benchmarking}
B.~Calli, A.~Walsman, A.~Singh, S.~Srinivasa, P.~Abbeel, and A.~M. Dollar,
  ``Benchmarking in manipulation research: The {YCB} object and model set and
  benchmarking protocols,'' \emph{Proceedings of the 2015 IEEE International
  Conference on Advanced Robotics (ICAR)}, 2015.

\bibitem{johnson2017clevr}
J.~Johnson, B.~Hariharan, L.~Van Der~Maaten, L.~Fei-Fei, C.~Lawrence~Zitnick,
  and R.~Girshick, ``{CLEVR}: A diagnostic dataset for compositional language
  and elementary visual reasoning,'' in \emph{Proceedings of the IEEE
  Conference on Computer Vision and Pattern Recognition (CVPR)}, 2017, pp.
  2901--2910.

\bibitem{kuhnle2017shapeworld}
A.~Kuhnle and A.~Copestake, ``{ShapeWorld} - {A} new test methodology for
  multimodal language understanding,'' \emph{arXiv preprint arXiv:1704.04517},
  2017.

\bibitem{mnih2014recurrent}
V.~Mnih, N.~Heess, A.~Graves, and K.~Kavukcuoglu, ``Recurrent models of visual
  attention,'' in \emph{Proceedings of the International Conference on Neural
  Information Processing Systems (NeurIPS)}, 2014, p. 2204–2212.

\bibitem{he2016deep}
K.~He, X.~Zhang, S.~Ren, and J.~Sun, ``Deep residual learning for image
  recognition,'' in \emph{Proceedings of the IEEE Conference on Computer Vision
  and Pattern Recognition (CVPR)}, 2016, pp. 770--778.

\bibitem{russakovsky2015imagenet}
O.~Russakovsky, J.~Deng, H.~Su, J.~Krause, S.~Satheesh, S.~Ma, Z.~Huang,
  A.~Karpathy, A.~Khosla, M.~Bernstein, \emph{et~al.}, ``{ImageNet} large scale
  visual recognition challenge,'' \emph{International Journal of Computer
  Vision}, vol. 115, no.~3, pp. 211--252, 2015.

\bibitem{hudson2018compositional}
D.~A. Hudson and C.~D. Manning, ``Compositional attention networks for machine
  reasoning,'' in \emph{Proceedings of the International Conference on Learning
  Representations (ICLR)}, 2018.

\bibitem{williams1992simple}
R.~J. Williams, ``Simple statistical gradient-following algorithms for
  connectionist reinforcement learning,'' \emph{Machine Learning}, vol.~8, no.
  3-4, pp. 229--256, 1992.

\bibitem{ha2018world}
D.~Ha and J.~Schmidhuber, ``World models,'' \emph{arXiv preprint
  arXiv:1803.10122}, 2018.

\bibitem{hill_environmental_2020}
F.~Hill, A.~Lampinen, R.~Schneider, S.~Clark, M.~Botvinick, J.~L. McClelland,
  and A.~Santoro, ``Environmental drivers of systematicity and generalization
  in a situated agent,'' in \emph{Proceedings of the International Conference
  on Learning Representations (ICLR)}, 2020.

\bibitem{chaplot2020object}
D.~S. Chaplot, D.~Gandhi, A.~Gupta, and R.~Salakhutdinov, ``Object goal
  navigation using goal-oriented semantic exploration,'' in \emph{In Neural
  Information Processing Systems (NeurIPS)}, 2020.

\end{thebibliography}

\end{document}